\documentclass{article}
\usepackage{ijcai20}

\usepackage{times}

\usepackage{soul}
\usepackage{url}
\usepackage[hidelinks]{hyperref}
\usepackage[utf8]{inputenc}
\usepackage[small]{caption}
\usepackage{graphicx}
\usepackage{amssymb,amsmath}
\usepackage{amsthm}
\usepackage[margin=1in]{geometry}
\usepackage{floatflt}
\usepackage{booktabs}
\usepackage{algorithm}
\usepackage{algorithmic}

\title{A study of local optima for learning feature interactions using neural networks}

\author{
Yangzi Guo$^1$\footnote{Contact Author}\and
Adrian Barbu$^2$\\
\affiliations
$^1$Mathematics Department, Florida State University\\
$^2$Statistics Department, Florida State University\\
\emails
yguo@math.fsu.edu,
abarbu@stat.fsu.edu
}

\DeclareGraphicsExtensions{.jpg,.pdf,.mps,.png}

\def\RR{\mathbb R}

\newcommand{\bb}{\mathbf{b}}

\newcommand{\bx}{\mathbf{x}}

\newcommand{\bw}{\mathbf{w}}

\newcommand{\bbeta}{{\boldsymbol{\beta}}}

\begin{document}

\maketitle

\begin{abstract}
In many fields such as bioinformatics, high energy physics, power distribution, etc., it is desirable to learn non-linear models where a small number of variables are selected and the interaction between them is explicitly modeled to predict the response.
In principle, neural networks (NNs) could accomplish this task since they can model non-linear feature interactions very well. However, NNs require large amounts of training data to have a good generalization.
In this paper we study the data-starved regime where a NN is trained on a relatively small amount of training data. For that purpose we study feature selection for NNs, which is known to improve generalization for linear models. 
As an extreme case of data with feature selection and feature interactions we study the XOR-like data with irrelevant variables. We experimentally observed
that the cross-entropy loss function on XOR-like data has many non-equivalent local optima, and the number of local optima grows exponentially with the number of irrelevant variables. 
To deal with the local minima and for feature selection we propose a node pruning and feature selection algorithm that improves the capability of NNs to find better local minima even when there are irrelevant variables. 
Finally, we show that the performance of a NN on real datasets can be improved using pruning, obtaining compact networks on a small number of features, with good prediction and interpretability.
\vspace{-3mm}
\end{abstract}

\vspace{-4mm}
\section{Introduction}

Many fields of science such as bioinformatics, high energy physics, power distribution, etc., deal with tabular data with the rows representing the observations and the columns representing the features (measurements) for each observation. 
In some cases we are interested in predictive models to best predict another variable of interest (e.g. catastrophic power failures of the energy grid). 
In other cases we are interested in finding what features are involved in predicting the response (e.g. what genes are relevant in predicting a certain type of cancer) and the predictive power is secondary to the simplicity of explanation. Furthermore, in most of these cases a linear model is not sufficient since the variables have high degrees of interaction in obtaining the response.

Neural networks (NN) have been used in most of these cases, because they can model complex interactions between variables, however they require large amounts of training data. 
We are interested in cases when the available data is limited and the NNs are prone to overfitting.

To get insight on how to train NNs to deal with such data, we will study the XOR data, which has feature interactions and many irrelevant variables.
The feature interactions are hard to detect in this data because they are not visible in any marginal statistics.

We will see the the loss function has many local minima that are not equivalent and that irrelevant features make the optimization harder when data is limited.
To address these issues we propose a node pruning and feature selection algorithm that can obtain a compact NN on a small number of features, thus helping deal with the case of limited data and irrelevant features.

\vspace{-2mm}
\subsection{Related Work}

\noindent {\bf Local minima. } 
Recent studies \cite{draxler2018essentially,garipov2018loss} have shown that the local minima of some convolutional neural networks
are equivalent in the sense that they have the same loss (energy) value and a path can be found between the local minima along which the energy stays the same. 
For this reason, we will focus our attention to fully connected neural networks and find examples where the local minima have different loss values.
Moreover, \cite{soudry2016no} proves that all differentiable local minima are global minima for the
one hidden layer NNs with piecewise linear activation and square loss. However, nothing is proved for non-differentiable local minima.
\begin{figure*}[t]
\centering
\includegraphics[width=5.cm]{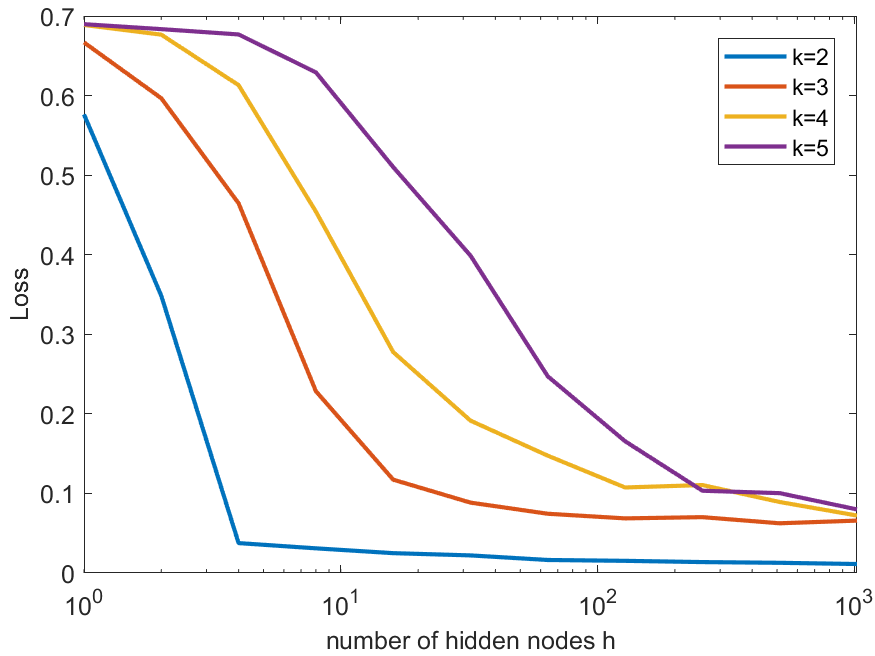}
\includegraphics[width=5.cm]{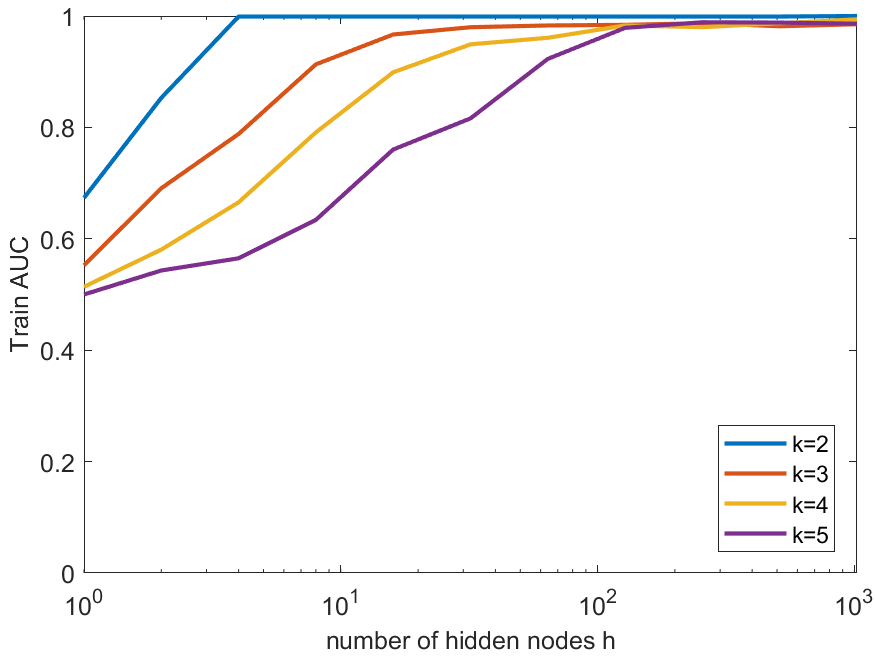}
\includegraphics[width=5.cm]{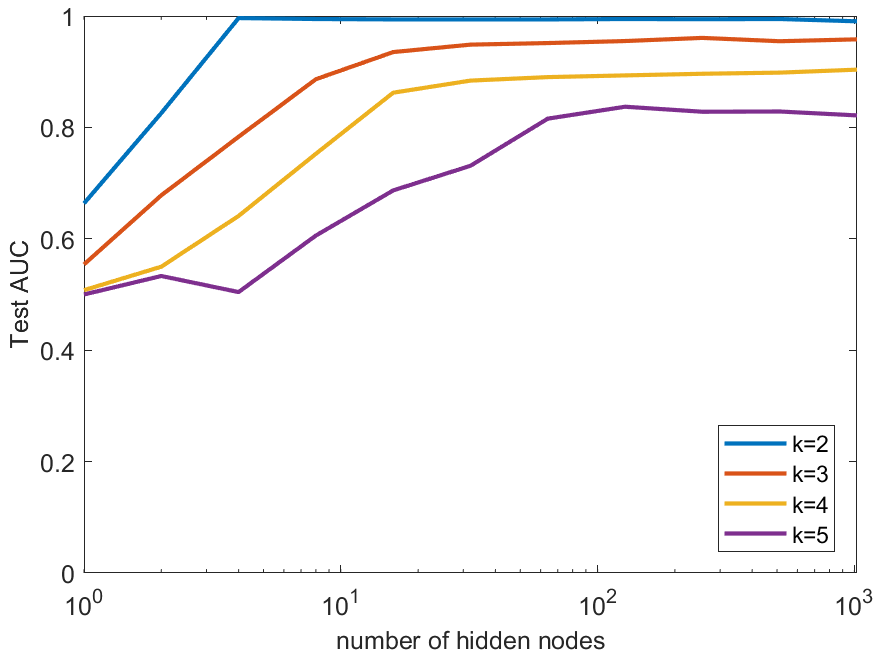}
\vskip -3mm
\caption{Loss value and training and test AUC vs number $h$ of hidden nodes for XOR data with no irrelevant variables, $n=1000$. 
}\label{fig:auctrue}
\vspace{-6mm}
\end{figure*}

\noindent {\bf Network pruning.} 
There has been quite a lot of work recently about neural network pruning, either for the purpose of improving speed and reducing complexity or giving insights about explaining the essential capability of the pruning technique. 
\cite{han2015learning} and \cite{han2015deep} propose the "Deep Compression", a three-stage technique, which significantly reduces the storage requirement for training deep neural networks without affecting their accuracy. \cite{liu2018rethinking} shows that for structured pruning methods, directly training the small target sub-network or pruned model with random initialization can achieve a comparable or even better performance than retraining using the remaining parameters after pruning. They also obtain similar results towards to a unstructured pruning method \cite{han2015learning} after fine-tuning the pruned sub-network on small-scale datasets. \cite{frankle2018lottery} introduces the Lottery Tickets Hypothesis which claims that a random-initialized dense neural network contains a sub-network that can be trained in isolation with the corresponding original initialized parameters to obtain the same test accuracy of the original network after training for the same number of iterations.
\vspace{-2mm}
\section{An Empirical Study of the Trainability of Data-Starved Neural Networks}

To study feature selection methods for neural networks, we will look at a challenging case study, the XOR problem with irrelevant variables. 
The $k$-dimensional XOR is a binary classification problem that can be formulated as
\vspace{-3mm}
\begin{equation}
  y(\bx)=I(\prod_{i=1}^k x_i>0), \forall \bx \in \RR^p \label{eq:xor}
\vspace{-3mm}  
\end{equation}
Observe that in this formulation the XOR data is $p$ dimensional but the degree of interaction is $k$-dimensional, with $k\leq p$. We call this data the $k$-D XOR in $p$ dimensions. In this paper we will work with $k\in \{3,4,5\}$, as $k=2$ is very simple. 
We assume that $\bx\in \RR^p$ is sampled uniformly from $[-1,1]^p$. The first $k$ features are the only ones used in generating the response, and we call them the {\em true features}.

The XOR problem an example of data that can only be modeled by using higher order feature interactions, and for which lower order marginal models have no discrimination power. This makes it very difficult to detect what features are relevant for predicting the response $y$.

The neural networks (NN) that we will study are two layer neural networks with ReLU activation for the hidden layer. These NNs can model the XOR data very well given sufficiently many hidden nodes. The networks will be trained using the Adam optimizer and the cross-entropy loss function.

To take the data variability out of the picture, for each $k$ we will construct a large dataset of $\bx$ values with a large enough number $N$ of observations and $P$ features and use subsets consisting of the first $n\leq N$ observations and $p\leq P$ features for our experiments. For each observation the $y$ is obtained deterministically using Eq. \eqref{eq:xor}.
The same way we construct a separate test set with $n=3000$ observations.

\vspace{-2mm}
\subsection{Deep local minima based on the true features}

In this section we study the NNs only on the true features used in generating the response, thus the feature selection is assumed given by an oracle. 
We study the local optima of the loss function that the NNs can obtain by training from a random initialization, for different numbers of hidden nodes. 
We are also interested in the connection between the number of hidden nodes $h$ and the training and test AUC. 

Since for each $k$ the dataset is assumed fixed, we will use the best test AUC obtained for each $h$ as a target that we would like to reach for the same $h$ even for data that has many irrelevant variables. 
 
\noindent {\bf Dependence on $h$.} In a first experiment, we train a NN with different numbers of hidden nodes $h$ on a dataset with $n=1000$ observations and 10 random initializations. Then for each $h$ we select the result with smallest loss out of the 10 initializations and compute its train and test AUC.
In Figure \ref{fig:auctrue} are shown the obtained values of the loss, train and test AUC vs number of hidden nodes $h$. We see that the loss decreases considerably first, then it stabilizes. Same happens with the train and test AUC. 
These experiments were used to select the number of hidden nodes that would obtain a maximal test AUC. At the same time the train AUC is larger than 0.95. The selected number of hidden nodes for each $k$ is shown in Table \ref{tab:hbest}. 

\begin{table}[ht]
\vspace{-3mm}
\centering
\begin{tabular}{lcccc}
\hline
Dataset size \  &$k=2$  & $k=3$ &$k=4$ &$k=5$ \\
\hline
$n=1000$   &4    &64 &128 &128  \\
\hline
\end{tabular}
\vskip -3mm
\caption{Number of hidden nodes $h$ for each $k$, for $p=k\in\{2,3,4,5\}$.}
\label{tab:hbest}
\vspace{-6mm}
\end{table}

\begin{figure*}[htb]
\centering
\includegraphics[width=5.cm]{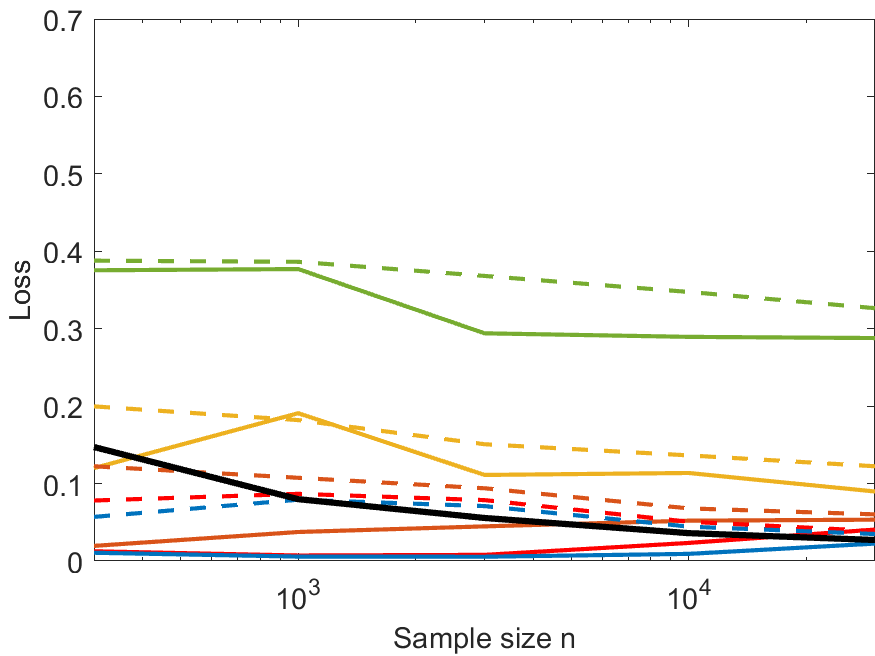}
\includegraphics[width=5.cm]{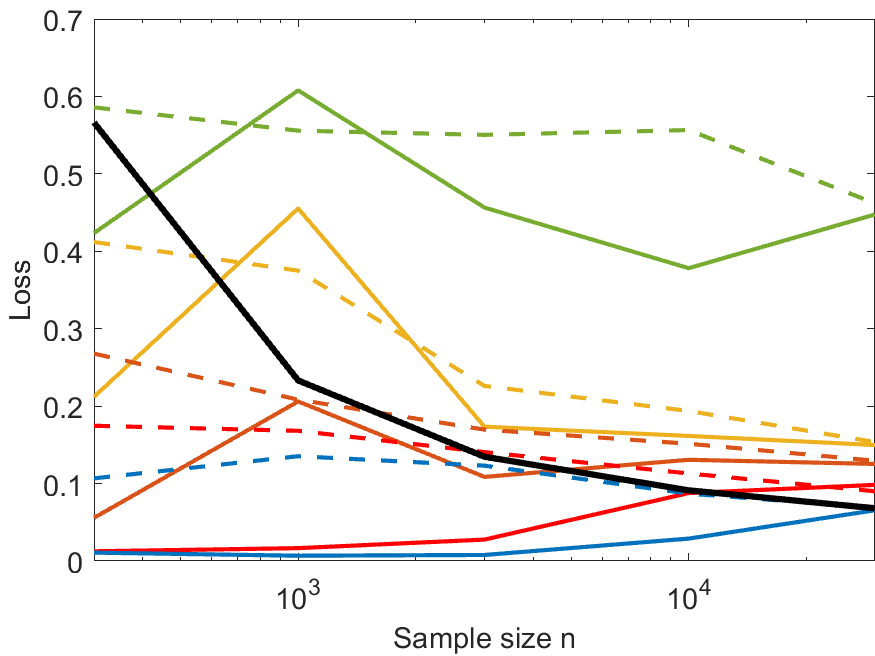}
\includegraphics[width=5.cm]{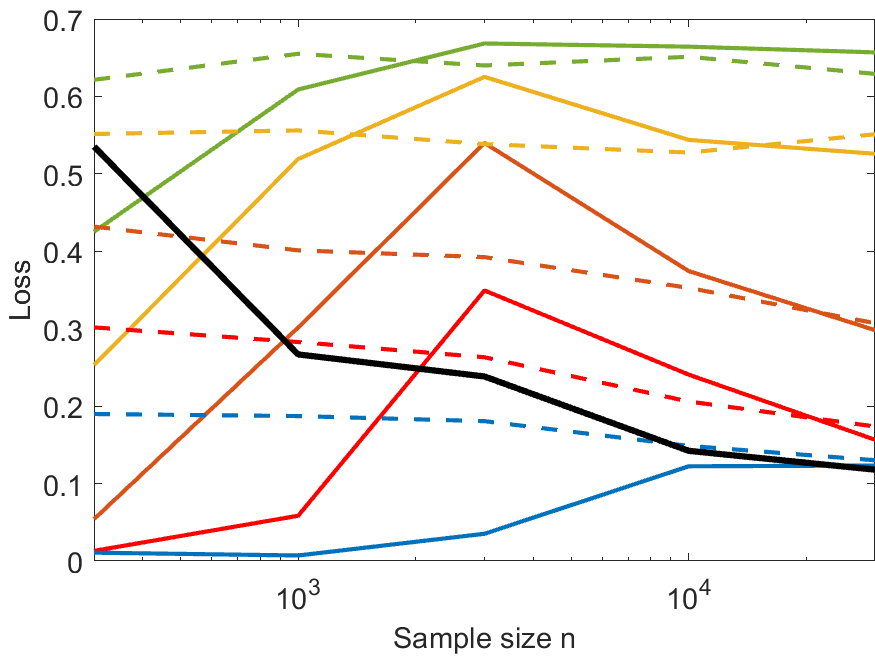}
\includegraphics[width=5.cm]{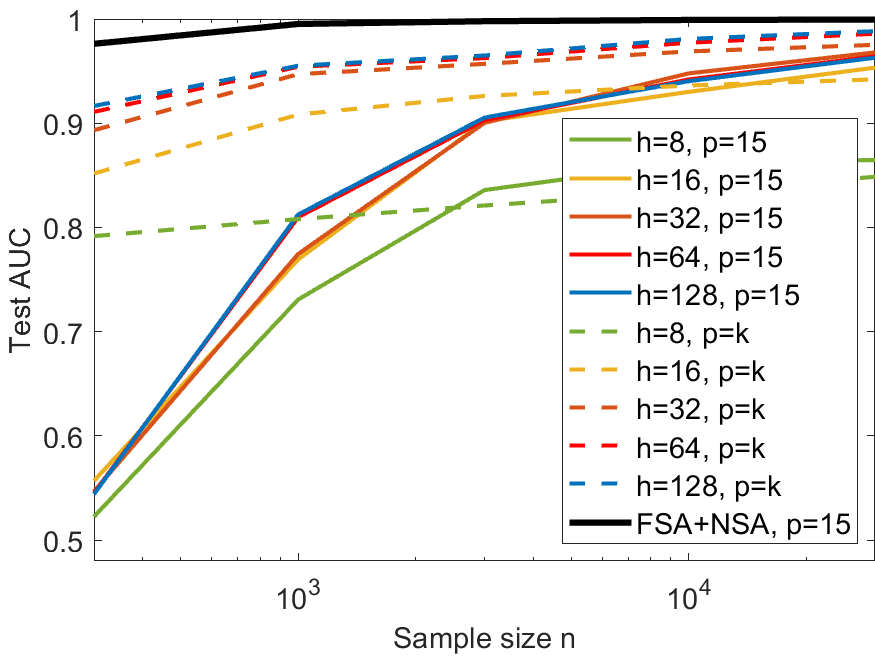}
\includegraphics[width=5.cm]{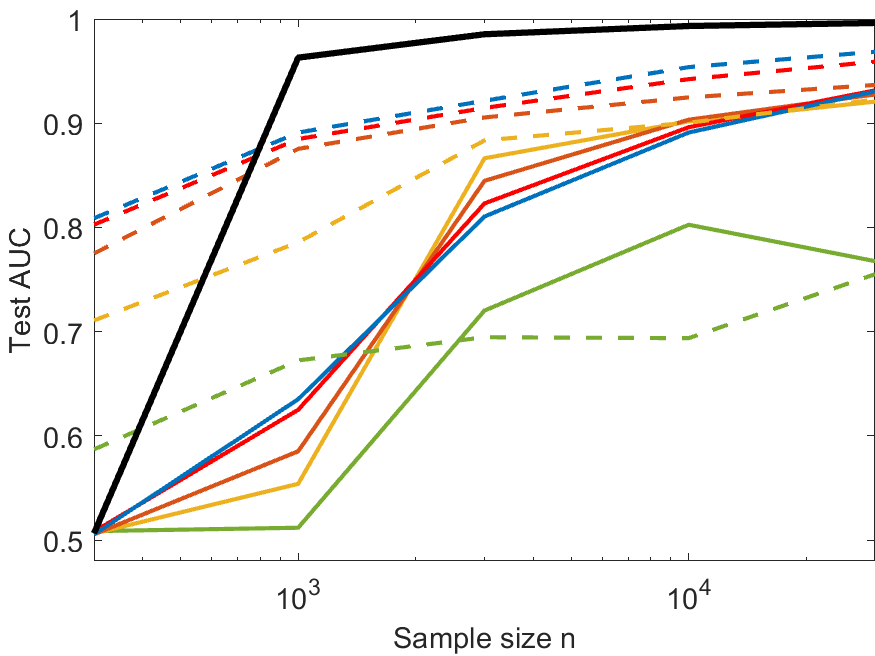}
\includegraphics[width=5.cm]{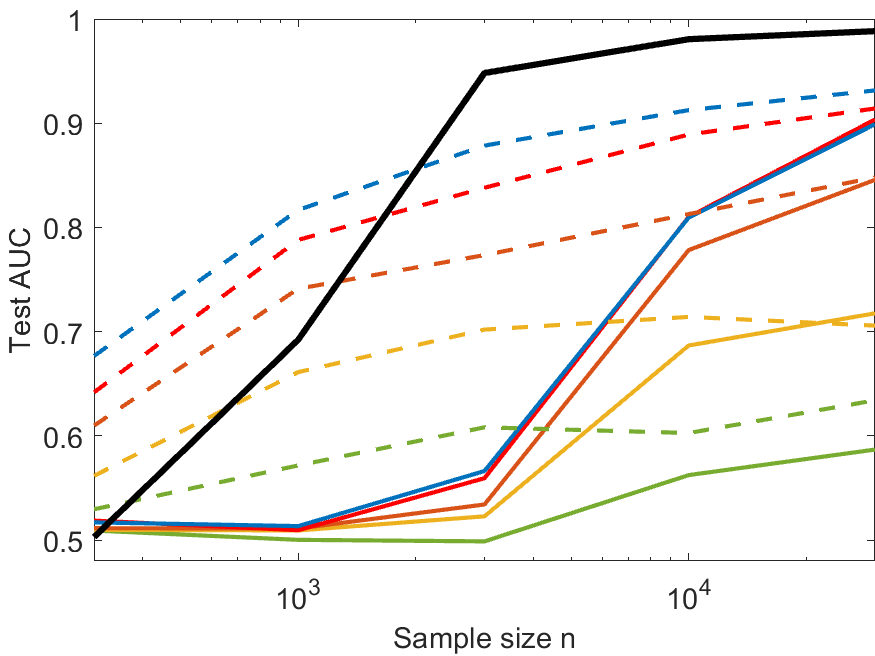}
\vskip -3mm
\caption{Training loss (top), and test AUC (bottom)  vs $n$ for $k=3$(left), $k=4$ (middle) and $k=5$ (right) for NNs with different number of hidden nodes $h$. Also shown is the result of the FSA+NSA procedure described in Section \ref{sec:nsafsa}.}\label{fig:loss_n}
\vspace{-5mm}
\end{figure*} 

\subsection{Local minima in training neural networks}
\noindent{\bf Loss values vs. $n$.} In Figure \ref{fig:loss_n} are shown the average loss values obtained from 10 random initializations vs sample size $n$ for different number of hidden nodes $h$. Shown are the loss values for data with $p=k$ (dashed lines) and for data with $p=15$ (solid lines), which has at least 10 irrelevant variables.
One can see that when there are no irrelevant variables ($p=k$), the obtained loss stays relatively constant and only slightly decreases with sample size. However, when there are irrelevant variables ($p=15$) the loss gradually increases, a sign of overfitting for small sample sizes, which could be addressed by variable selection. Moreover, the loss has a region where it takes large values ($n=1000$ for k=4 and $n=3000$ for k=5) for some network sizes $h$, which is a sign that the optimization is difficult there.
Looking at the test AUC, we see that it increases with the sample size, and for the data with irrelevant variables ( $p=15$) it never reaches the values of the test AUC for $p=k$, i.e. when we train a model on the $k$ relevant variables only.

\begin{figure*}[htb]
\centering
\includegraphics[width=5.cm]{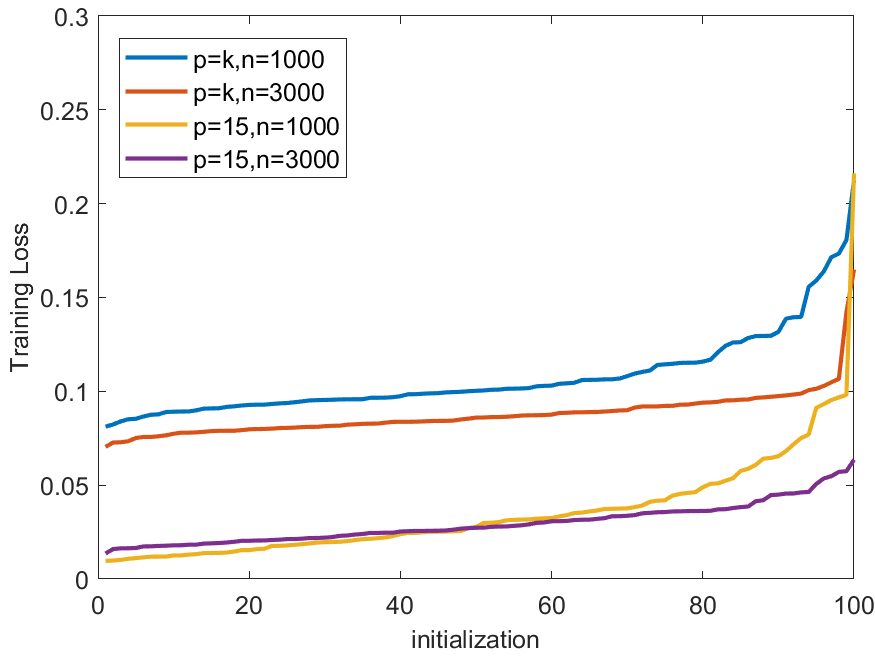}
\includegraphics[width=5.cm]{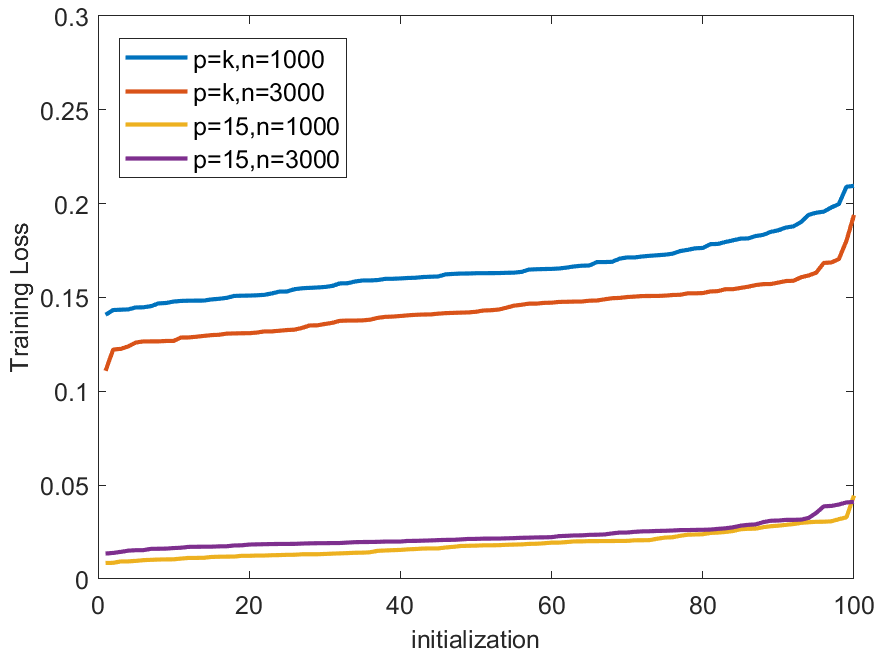}
\includegraphics[width=5.cm]{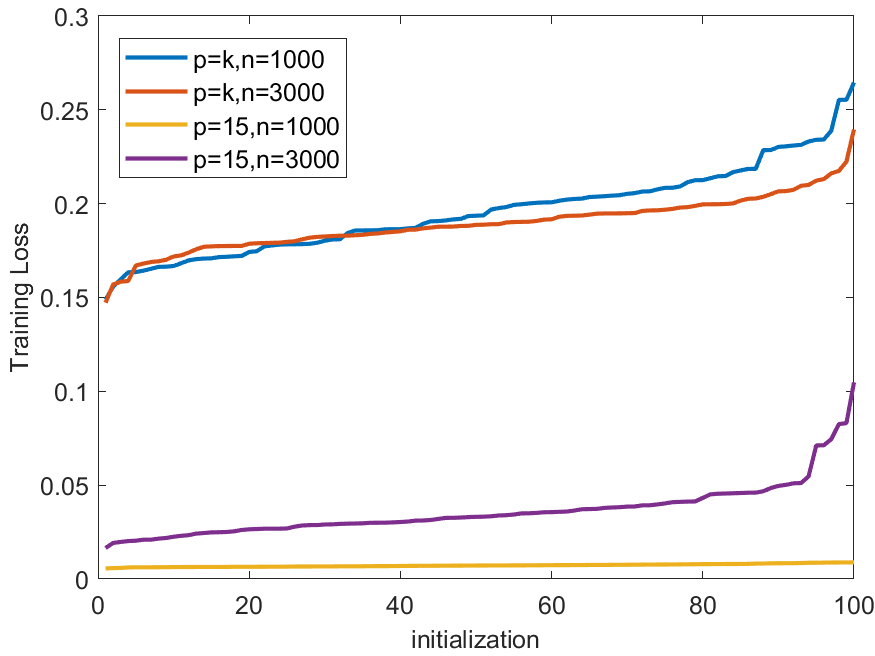}
\includegraphics[width=5.cm]{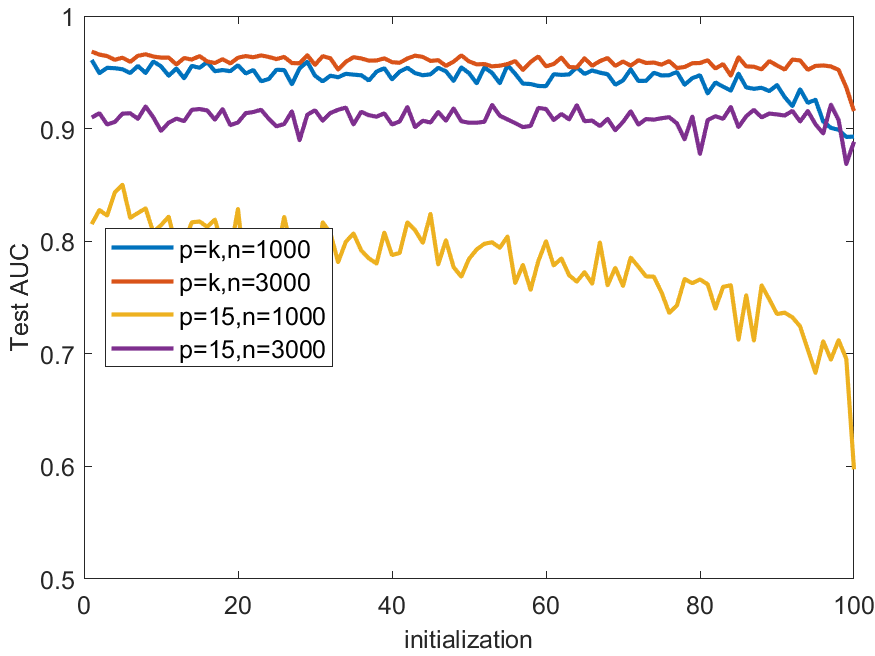}
\includegraphics[width=5.cm]{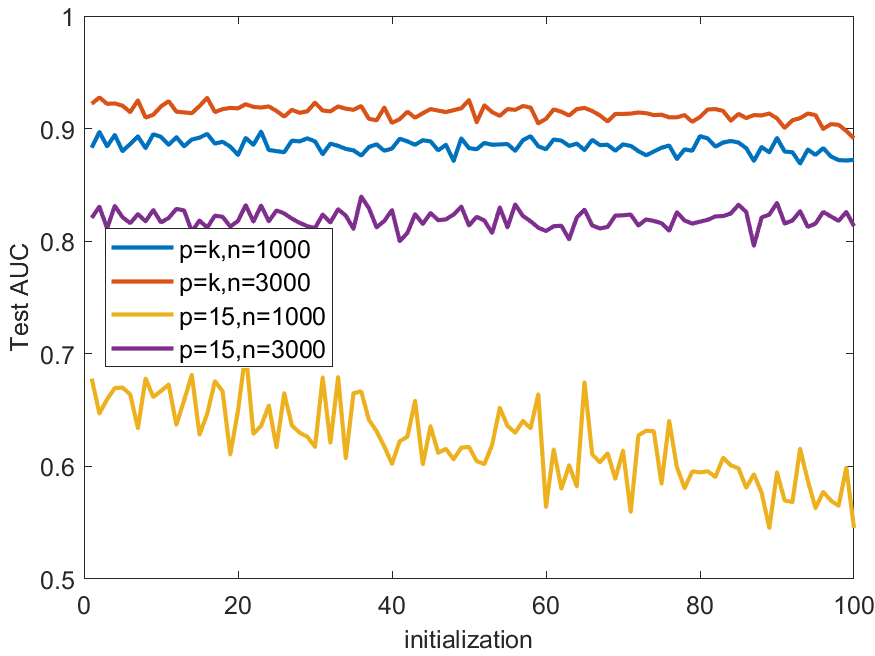}
\includegraphics[width=5.cm]{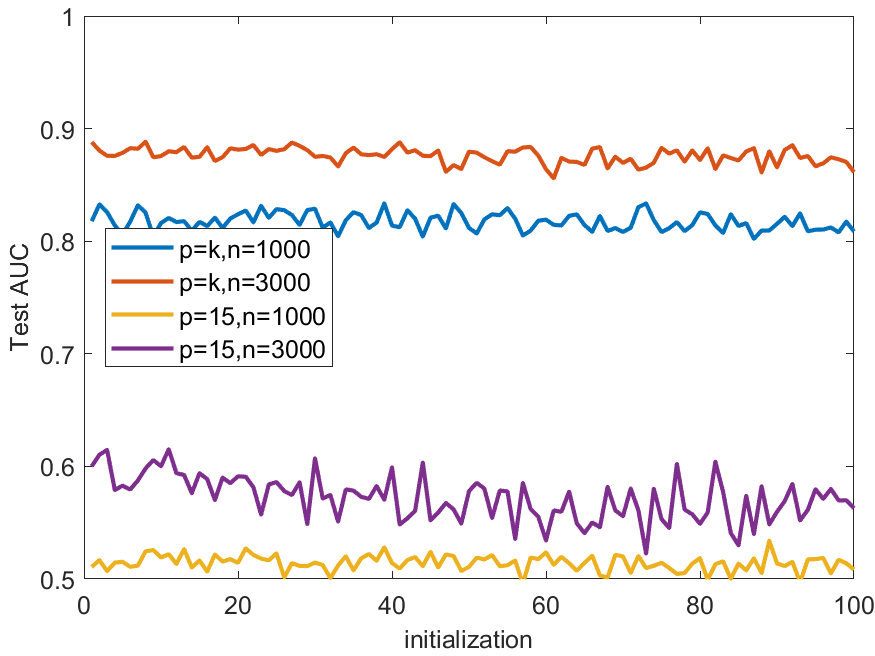}
\vskip -3mm
\caption{Training loss (top) and test AUC (bottom) for 100 random initializations, $k=3$ (left), $k=4$ (middle) and $k=5$ (right). The initializations have been sorted by the training loss value. The number of hidden nodes is given in Table \ref{tab:hbest}. }\label{fig:loss}
\vspace{-5mm}
\end{figure*} 
\noindent{\bf Local minima.} To study the local minima of a NN on the XOR data, we trained a NN with 100 random initializations. The number of hidden nodes was taken from Table \ref{tab:hbest}.

\begin{floatingfigure}[r]{4.5cm}
\vspace{-4mm}
\centering
\includegraphics[width=4.5cm]{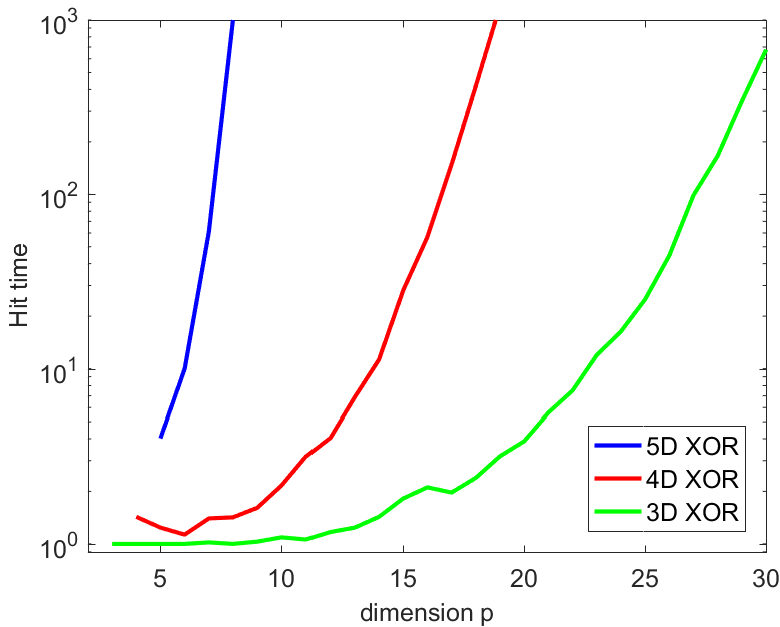}
\vskip -3mm
\caption{Hit time vs dimension $p$ for different XOR problems with $n=3000$ observations.}\label{fig:hit}
\vspace{-2mm}
\end{floatingfigure}The local minima were sorted by the loss value and their loss, train and test AUC are shown in Figure \ref{fig:loss}. We see that the loss values are clearly different and they reflect in different training and test AUCs. 
Since the dataset is the same for all initializations, the fact that the loss values are different indicates that there are many local minima with different values.

\noindent{\bf Hit time.} To see how hard to find are the local minima, we compute the hit time, which we define as the average number of random initializations required to find a local minimum with a train AUC of at least 0.95. 
The hit time is displayed in Figure \ref{fig:hit} for NNs with 20 hidden nodes and $n=3000$ observations.
Observe that the hit time quickly blows up as $p$ increases and has a super-exponential dependence on $p$. It is impractical to learn NN models on 3D, 4D or 5D XOR data when there are hundreds of irrelevant variables.
\begin{figure*}[htb]
\centering
\includegraphics[width=5.cm]{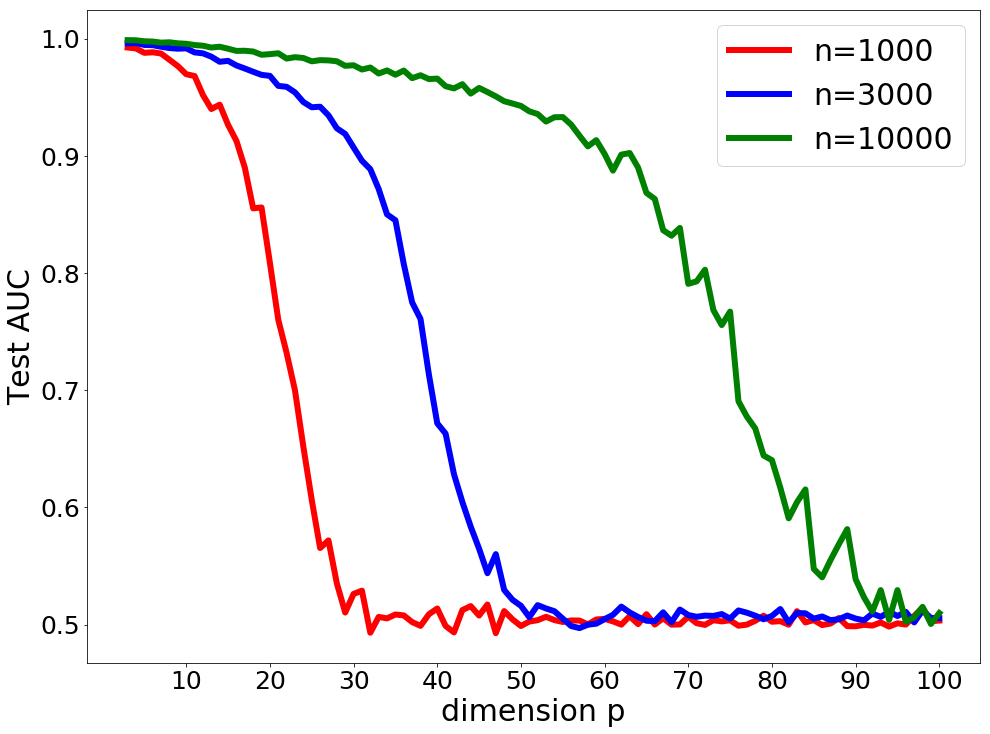}
\includegraphics[width=5.cm]{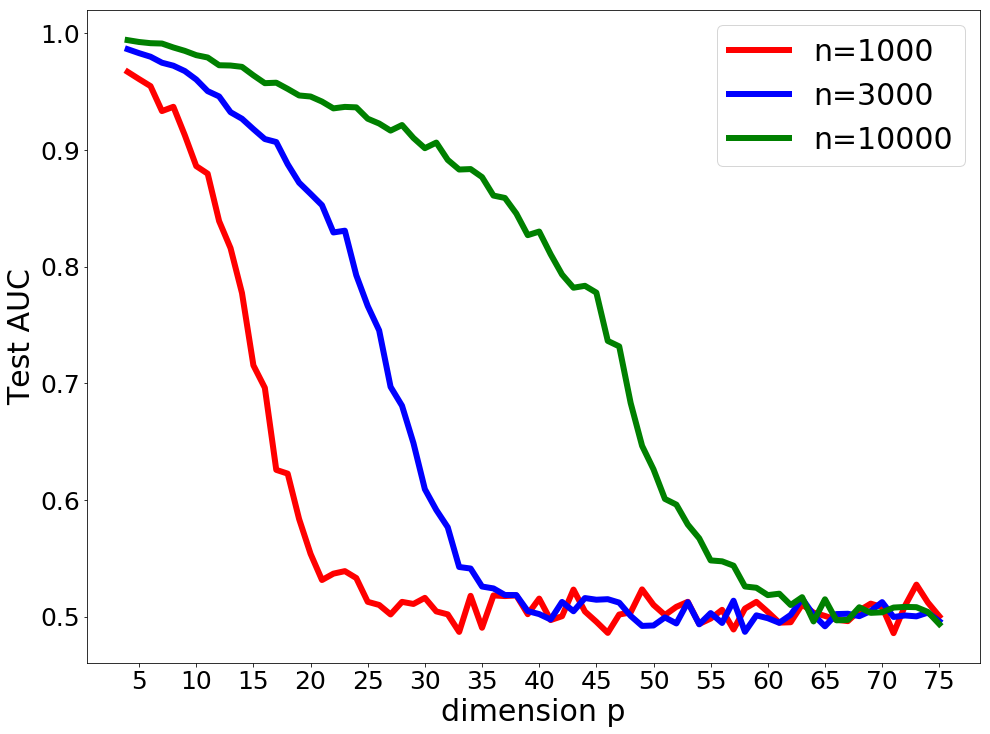}
\includegraphics[width=5.cm]{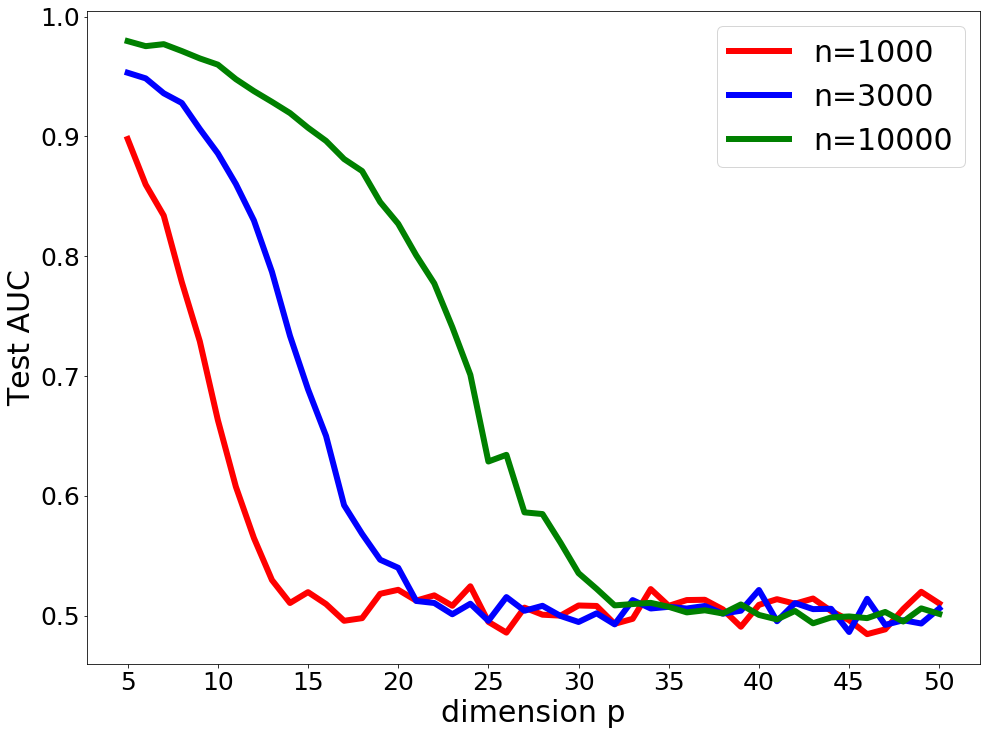}
\vskip -3mm
\caption{Test AUC of best loss minimum out of 10 random initializations vs. data dimension $p$ for a NN with 500 hidden nodes. Left: $k=3$. Middle: $k=4$. Right: $k=5$.}\label{fig:bestauc}
\vspace{-5mm}
\end{figure*} 
\noindent {\bf Dependence on $p$.} As we see from previous observations, the NNs can handle the XOR data if $p$ is small, but even if in the range NN can work, the test AUC still decreases as $p$ increases. 
We also observe that increasing the number of hidden node in NNs may not be very helpful for improving the test AUC. To demonstrate this observation, for different numbers of hidden node, we train a NN with 10 random initializations and keep the best test AUC and its associated training AUC among the 10 trials. 
We repeat this process 10 times and display Figure \ref{fig:nodeauc} the average test and train AUC vs the number of hidden nodes. When the number of hidden nodes increases, the training AUC becomes better and better and finally it reaches 1.0. 
But the test AUC are a different story, it quickly reaches its best value when the number of hidden nodes is relatively small, and then no further improvement happens as the number increases. 
This tells us that increasing the number of hidden nodes will make too many irrelevant hidden nodes exist in the NN, and lead to overfitting.

From this empirical study we conclude:
\begin{itemize}
\item If the training data is difficult (such as the XOR data), not all local minima are equivalent, since in Figure \ref{fig:loss} there was a large difference between the largest and smallest loss values as well as the corresponding test AUCs.
\item For a fixed $p$ the optimization problem is harder for data starved NNs, when the sample size $n$ is in a certain range, but not large enough.
\item For a fixed training size $n$, the number of shallow local minima quickly blows up as the number of irrelevant variables increases and finding the deep local minima becomes extremely hard.
\item If the number of irrelevant variables is not too large (e.g. $p=15$ as in Figure \ref{fig:loss_n}), an NN with a sufficiently large number of hidden nodes will find a deep optimum more often than one with a small number of hidden nodes, but it might overfit.
\end{itemize}
These conclusions are the basis for the proposed pruning methodology presented in the next section.

\section{Node and Feature Selection for Neural Networks}\label{sec:nsafsa}

The above study showed how important it is to remove the irrelevant variables when training neural networks on difficult data with a small number of observations.

We use neural networks with one hidden layer and ReLU activation for the hidden layer. If the hidden layer has $h$ neurons and the input $\bx\in\RR^p$, we can represent the weights of the hidden nodes as vectors $\bw_j\in \RR^p,  j=1,...,h$, the biases as a vector $\bb=(b_1,...,b_h)\in \RR^h$ and the weights of the output neuron as a vector $\bbeta=(\beta_1,...,\beta_h)\in \RR^h$. Denoting the ReLU activation as $\sigma(x)=\max(0,x)$ we can write the neural network as:
\vspace{-2mm}
\begin{equation}
f(\bx)=\sum_{j=1}^h \beta_j\sigma(\bw_j^T \bx+b_j)
\vspace{-1mm}
\end{equation}

\subsection{Node Selection with Annealing for NN} \label{sec:nsa}

To find better local optima, we propose to start with a NN with many hidden neurons and use a pruning method similar to the Feature Selection with Annealing \cite{barbu2017feature} to select the well trained hidden nodes and remove the rest. 
\vspace{-3mm}
\begin{algorithm}[htb]
	\caption{{\bf Node Selection with Annealing (NSA)}}
	\label{alg:nsa}
	\begin{algorithmic}
		\STATE {\bfseries Input:} Training set $T=\{(\bx_i,y_i)\in \RR^p\times \RR\}_{i=1}^{n}$, desired number $h$ of hidden neurons, starting number $H$ of hidden neurons, annealing schedule $M_e,e=1,..,N^{iter}$.
		\STATE {\bfseries Output:} Trained NN with $h$ hidden neurons.
	\end{algorithmic}
	\begin{algorithmic} [1]
		\STATE Initialize a NN with $H$ hidden neurons with random initialization
		\FOR {$e = 1$ to $N^{iter}$}
			\STATE  Train the NN for 1 epoch
			\FOR {$j= 1$ to $h$}
			\STATE Normalize hidden node $j$:
\vspace{-2mm}
			\begin {equation} \label{eq:nodenorm}
			 \beta_j\hspace{-0.5mm}\leftarrow \hspace{-0.5mm}  \|\bw_j\|\beta_j, 
			 b_j\hspace{-0.5mm}\leftarrow\hspace{-0.5mm} \frac{b_j}{\|\bw_j\|}, 
			 \bw_j\hspace{-0.5mm}\leftarrow \hspace{-0.5mm}\frac{\bw_j}{\|\bw_j\|}
\vspace{-4mm}
			  \end{equation}
			\ENDFOR
			\STATE Remove hidden nodes to keep the $h_e$ nodes with largest $|\beta_j|$
		\ENDFOR
\end{algorithmic}
\end{algorithm}
\vspace{-4mm}

However, the NN has some built-in redundancy that we need to take into consideration when comparing the hidden nodes with each other. 
Observe that if we multiply $\bw_j$ and $b_j$ by a constant $c>0$ and divide $\beta_j$ by the same $c$ we obtain an equivalent NN that has exactly the same output, due to the fact that we use ReLU activation. 
We can remove this redundancy and normalize the hidden neurons by normalizing their weight vectors $\bw_j$. 
The proposed method for pruning the nodes including  this normalization step is presented in Algorithm \ref{alg:nsa}.

The node annealing schedule $h_e$ follows the equation:
\vspace{-2mm}
\[
h_e=h + [(H-h)\max(0,\frac{N_1-2e_1}{2e_1\mu+N_1})]
\vspace{-2mm}
\]
\begin{floatingfigure}[r]{4.5cm}
\vspace{-2mm}
\centering
\includegraphics[width=4.5cm]{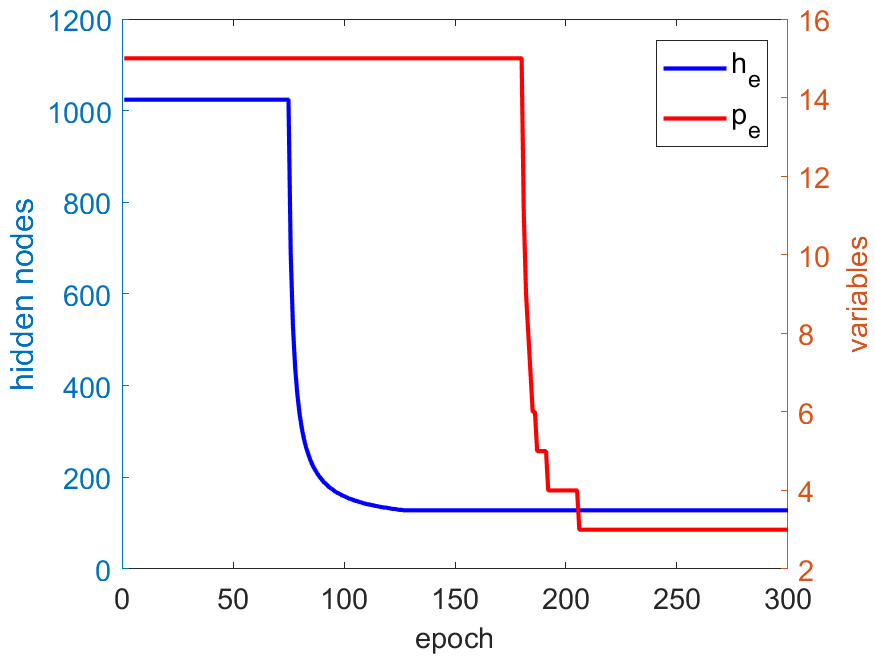}
\vskip -4mm
\caption{Annealing schedules $h_e$ for NN nodes and $p_e$ for input variables used for $p=15,k=3$ data.}\label{fig:schedule}
\vspace{-5mm}
\end{floatingfigure}where $N_1=3N^{iter}/4$, $e_1=\max(0,e-N^{iter}/4)$, $H$ is the starting number of nodes (we used $H=1024$) and $h$ is the final number of nodes, e.g. $h=128$. The annealing parameter $\mu$ was set to $\mu=30$.
An example is shown as the blue curve in Figure \ref{fig:schedule}.

\subsection{Feature Selection for NN}


 We can use the node selection procedure from Section \ref{sec:nsa} to train better NNs than by random initialization when there are irrelevant variables. However, the irrelevant variables will still have a negative influence on the obtained model, and an even better model can be obtained by removing the irrelevant features.

After normalizing the NN using Eq. \eqref{eq:nodenorm}, we can compute the group weight (relevance) of each feature using the $L2$-norm of the corresponding variables in the $h$ weight vectors $\bw_j$:
\vspace{-3mm}
\begin{equation}
r_i^2= \sum_{j=1}^h w_{ji}^2 \label{eq:featnorm}
\vspace{-3mm}
\end{equation}



Using this group criterion we can use Feature Selection with Annealing to select the relevant features for a NN. The procedure is described in Alg. \ref{alg:fsa}.
\vspace{-3mm}
\begin{algorithm}[htb]
	\caption{{\bf Feature Selection with Annealing with NSA (FSA+NSA)}}
	\label{alg:fsa}
	\begin{algorithmic}
		\STATE {\bfseries Input:} Training set $T=\{(\bx_i,y_i)\in \RR^p\times \RR\}_{i=1}^{n}$, desired number $k$ of features, annealing schedule $M_e,e=1,..,N^{iter}$.
		\STATE {\bfseries Output:} Trained NN depending on exactly $k$ features.
	\end{algorithmic}
	\begin{algorithmic} [1]
		\STATE Train a NN using Algorithm \ref{alg:nsa}.
		\FOR {$e = 1$ to $N^{iter}$}
			\STATE  Train the NN for 1 epoch
			\STATE Normalize the hidden nodes using Eq. \eqref{eq:nodenorm}
			\STATE Compute the feature weights $r_j$ using Eq. \eqref{eq:featnorm}
			\STATE Keep the $p_e$ features with largest $r_j$
		\ENDFOR
\end{algorithmic}
\end{algorithm}
\vspace{-3mm}

The variable annealing schedule $p_e$ follows the equation:
\vspace{-2mm}
\[
p_e=k + [(p-k)\max(0,\frac{N_2-2e_2}{2e_2\mu+N_2})]
\vspace{-2mm}
\]
where $N_1=0.4N^{iter}$ and $e_2=\max(0,e-0.6N^{iter})$.
An example is shown as the red curve in Figure \ref{fig:schedule}.

\section{Experiments}

In this section we perform experiments on the XOR data and some real datasets.
All the experiments were trained with the Adam optimizer \cite{kingma2014adam} with the default learning rate 0.001 and weight decay 0.0001.

\subsection{XOR Data}

We ran FSA+NSA on the XOR data with $p=15$, for $N^{iter}=300$ epochs, where the node pruning happened after $N^{iter}/4=75$ epochs and feature selection started after $180$ epochs. We started with $H=1024$ nodes and pruned them to $h=128$.
The result is shown as the black curve in Figure \ref{fig:loss_n}.
One can see that the FSA+NSA procedure does a very good job in selecting the features and training a small model on the selected features. In most cases it even outperforms (in terms of test AUC) the NN model trained on the true features.
\begin{figure*}[htb]
\centering
\includegraphics[width=3.2cm]{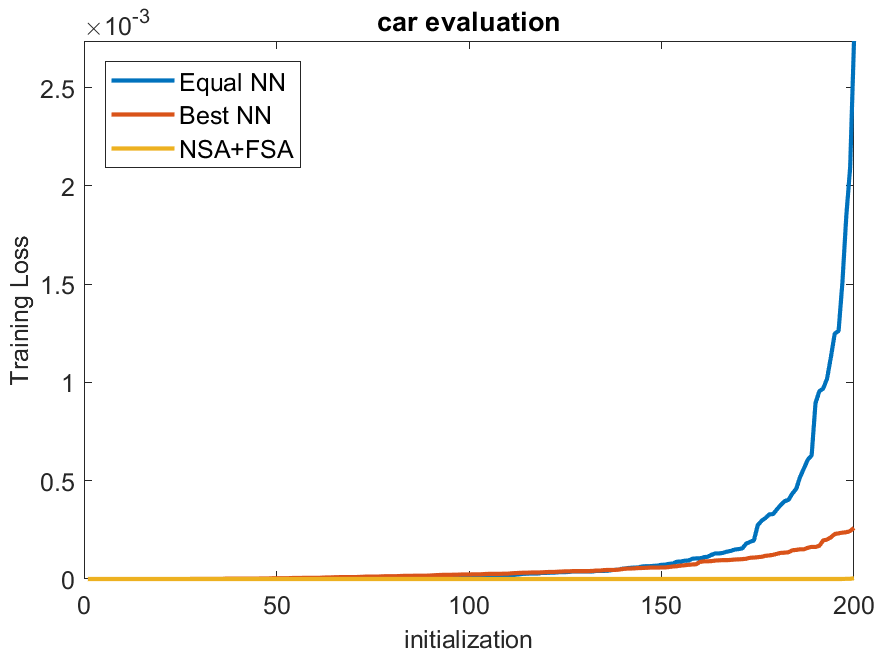}
\includegraphics[width=3.2cm]{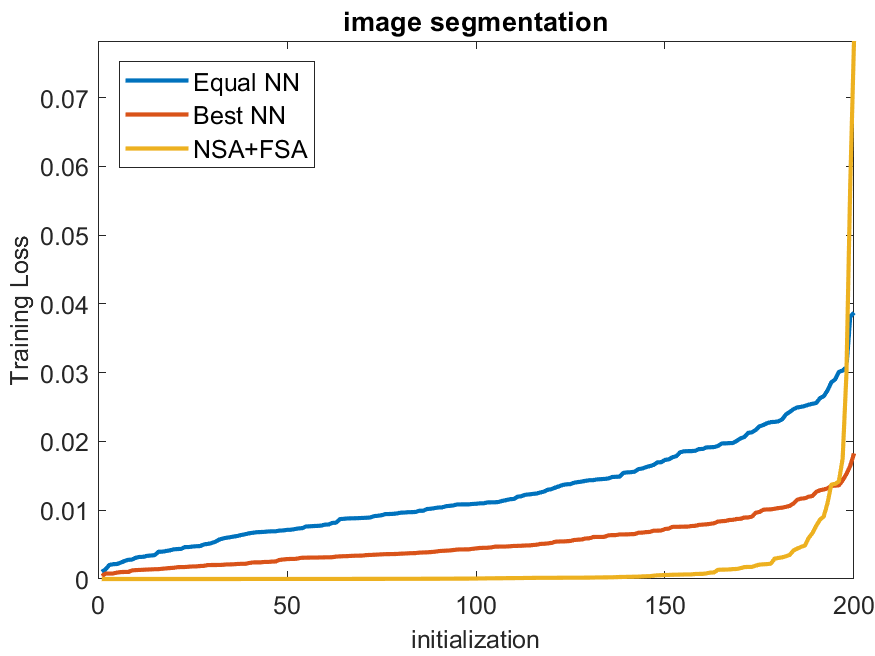}
\includegraphics[width=3.2cm]{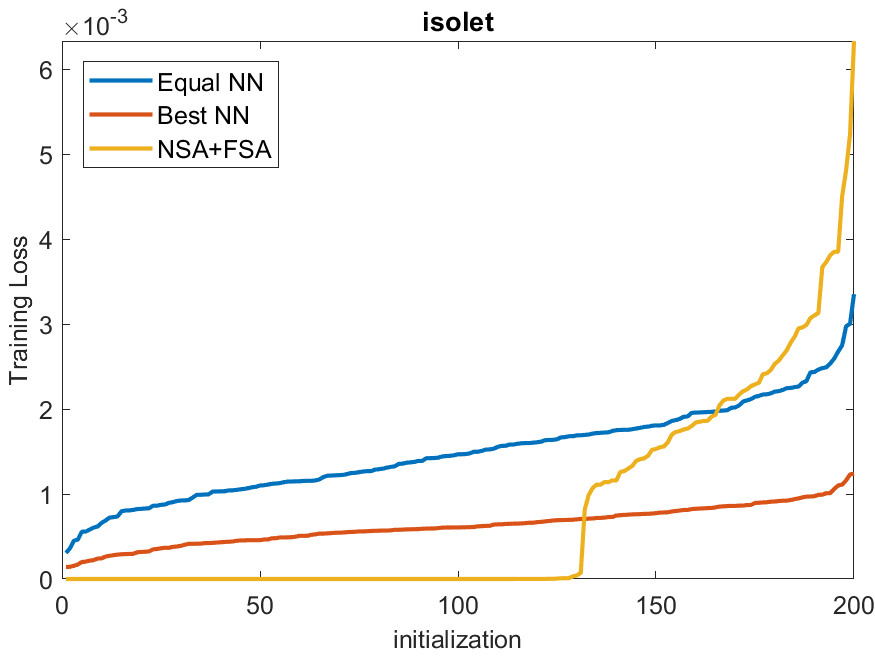}
\includegraphics[width=3.2cm]{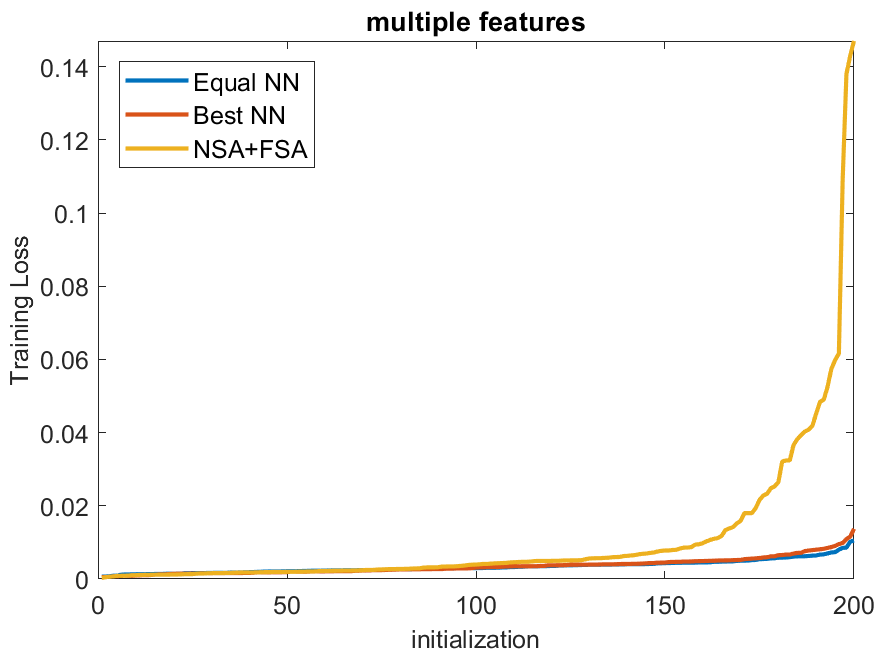}
\includegraphics[width=3.2cm]{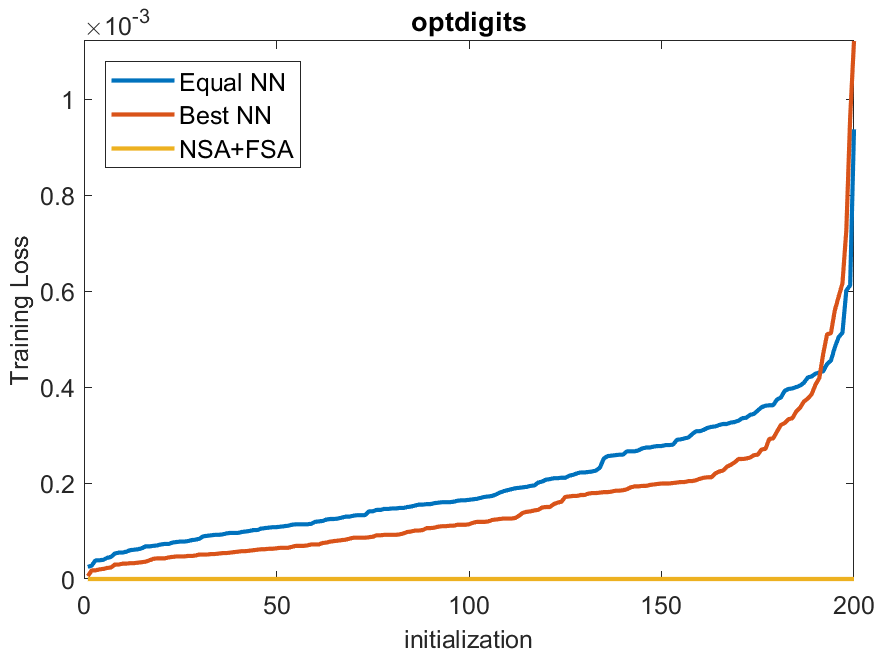}
\vskip -3mm
\caption{Sorted loss values for 200 initializations obtained on the five real datasets.}\label{fig:slossdata}
\vspace{-3mm}
\end{figure*}
\begin{table*}[ht]
\centering
\begin{tabular}{llll}
             & NN(best)            & NN(equivalent)   & FSA+NSA   \\
\hline
\multicolumn{4}{l}{ Car Evaluation, $p=21, n=1728$, $4$ classes.}            \\
\hline
Number of weights (nodes)     \  & 1600 (64) \  & 150 (6) \  & 120+32 = 152                 \\
Test Accuracy             \  & \bf{100.0$\pm$0.00}    \  & 98.23$\pm$0.06   \  & \bf{100.0$\pm$0.00}   \\
\hline
\multicolumn{4}{l}{ Image Segmentation, $p=19, n=2310$, $7$ classes.   }      \\
\hline
Number of weights (nodes) \  & 6656 (256) \  & 364 (14) \  & 266+98 = 364             \\
Test Accuracy             \  & 96.87$\pm$0.72    \  & 96.27$\pm$0.58   \  & \bf{98.40$\pm$0.32}        \\
\hline
\multicolumn{4}{l}{ Optical Recognition of Handwritten Digits, $p=64, n=5620$, $10$ classes.}       \\
\hline
Number of weights (nodes)     \  & 37888 (512) \  &1998 (27) \  & 1792+160 = 1952      \\
Test Accuracy             \  & 98.80$\pm$0.29    \  & 98.25$\pm$0.19   \  & \bf{99.01$\pm$0.20}   \\
\hline
\multicolumn{4}{l}{ Multiple Features, $p=216, n=2000$, $10$ classes.}         \\
\hline
Number of weights (nodes)    \  & 14464 (64) \  & 904 (4) \  & 583+320 = 903                 \\
Test Accuracy             \  & 97.85$\pm$0.80    \  & 95.45$\pm$0.98   \  & \bf{98.15$\pm$0.82}   \\
\hline
\multicolumn{4}{l}{ ISOLET, $p=617, n=7797$, $26$ classes.}                    \\
\hline
Number of weights (nodes)     \  &41152 (64) \  &5787 (9) \  & 4683+1118 = 5801                 \\
Test Accuracy             \  & 96.73$\pm$0.50    \  & 94.31$\pm$0.61   \  & \bf{96.91$\pm$0.54}   \\
\hline
\end{tabular}
\vspace{-3mm}
\caption{Performance results of NN(best), NN(equivalent) and FSA+NSA for each dataset.}
\label{tab:real_data_res}
\vspace{-6mm}
\end{table*}
\subsection{Real Datasets}

In this section, we perform an evaluation on a number of real multi-class datasets to compare the performance of a fully connected NN and the compact NN obtained by FSA+NSA. 
The real datasets were carefully selected from the UCI ML repository \cite{Dua:2019} to ensure that the dataset is not too large (the number of data points less than 10000) and that a standard fully connected neural network (with one hidden layer) can have a reasonable generalization power on this data. 
If a dataset is large, then the loss landscape is simple and  the neural network can be trained easily, so there is no need for pruning to escape bad optima.
If a dataset is such that a neural network can rarely be trained on it successfully, it means that the loss might not have any good local optima, then again pruning might not make sense. 

Our real dataset experiments are not aimed at comparing the performance with other classification techniques, but to test the effectiveness of FSA+NSA in guiding neural networks to find better local optima, we will combine all the samples including training, validation and testing data to form a single dataset for each data type first, and then divide them into a training and testing set with a ratio $4:1$. 
The obtained training dataset will be used in a 10-run averaged 5-fold cross-validation grid search training process to find the best hyper-parameter settings of a one hidden layer fully connected neural network. 
After getting the best hyper-parameter setting from the cross-validation, we use them to retrain the fully connected NNs with the entire training dataset 10 different times, and each time we record the best test accuracy. This  procedure is used for the fully connected NN, and the NN with FSA+NSA with different sparsity levels and record the best sparsity level and testing accuracy. 
Finally, we will also train a so-called "equivalent" fully connected neural network with roughly the same number of connections as the best sparse neural network we get from FSA+NSA. 

The number of hidden nodes was searched in $\{16,32,64,128,256,512\}$, the L2 regularization coefficient was searched in $\{0.0001,0.001,0.01,0.1\}$, the batch size was searched in $\{16,32,64\}$. Other NN training techniques like Dropout \cite{srivastava2014dropout} and Batch Normalization \cite{ioffe2015batch} were not used in our experiments due to the simplicity of the architecture of experimented NNs. 
The sorted loss values of the models with 200 random initializations are shown in Figure \ref{fig:slossdata}. 
The comparison results are listed in Table \ref{tab:real_data_res}.

We see from Table \ref{tab:real_data_res} that using FSA+NSA to guide the search for a local optimum leads to NNs with good generalization on all these datasets, easily outperforming a NN of an equivalent size (with a similar number of weights) and in most cases even the standard NN with the best generalization to unseen data. We see from Fig. \ref{fig:slossdata} that the FSA+NSA can obtain lower loss values than the other networks in all cases but one.

The experiments show that  the XOR data is indeed an extreme example where  deep local optima are be hard to find, but even these datasets exhibit some non-equivalent local optima and the things we learned from the XOR data carry over to these datasets to help us train NNs with better generalization.

\vspace{-3mm}
\section{Conclusion}

This paper presented an empirical study of the trainability of neural networks and the connection between the amount of training data and the loss landscape.
We observed that when the training data is large (where "large" depends on the problem), the loss landscape is simple and easy to train. When the training data is limited, the number of local optima can become very large, making the optimization problem very difficult. 
For these cases we introduce a method for training a neural network that avoids many local optima by starting with a large model with many hidden neurons and gradually removing neurons to obtain a compact network trained in a deep minimum. Moreover, the performance of the obtained pruned sub-network is hard to achieve by retraining  using random initialization, due to the existence of many shallow local optima around the deep minimum.
Experiments also show that the pruning method is useful in improving generalization on the XOR data and on a number of real datasets.

Many mature fields of science, such as physics, material science, electrical engineering etc., have two branches: one theoretical  and one experimental, and researchers are usually specialized on only one of these branches. 
The experimental scientists are skilled in designing and conducting experiments, handling different tools and devices and observing phenomena. These phenomena are later explained by their theoretical colleagues that are specialized in proving things theoretically or simulating them numerically. Sometimes  the opposite happens when a theoretical scientist predicts a certain phenomenon that is later verified by an experimentalist. Each branch requires different sets of skills and there are very few scientists in those fields that are both theoretical and experimental.

We feel that Machine Learning has reached a degree of maturity where it could also benefit from such a division. Some  studies could be purely experimental and leave the theoretical justification to other more theoretically skilled researchers. In this regard, our paper is a purely experimental study, observing some phenomena and providing some intuitive solutions. We leave the theoretical study of the phenomena as well as the proof of the theoretical grounding for our FSA+NSA algorithm as future work for somebody with the appropriate skill set. 

\bibliography{references}
\bibliographystyle{ijcai20}

\end{document}